\documentclass[11pt,a4paper]{article}
\usepackage{acl2020}
\usepackage{times}
\usepackage{latexsym}
\hypersetup{draft}

\usepackage{microtype}

\aclfinalcopy 
\def\aclpaperid{3282} 

\usepackage{amsmath,amsfonts,bm}

\def\eqref#1{equation~\ref{#1}}

\def\1{\bm{1}}

\DeclareMathAlphabet{\mathsfit}{\encodingdefault}{\sfdefault}{m}{sl}
\SetMathAlphabet{\mathsfit}{bold}{\encodingdefault}{\sfdefault}{bx}{n}

\usepackage{url}
\usepackage{epsfig,endnotes,graphicx,subcaption,xspace}
\usepackage{wrapfig}
\usepackage{diagbox}
\usepackage{multirow}
\usepackage{booktabs}

\newcommand{\sys}{DeFormer}

\newcommand{\eat}[1]{}

\title{DeFormer: Decomposing Pre-trained Transformers \\for Faster Question Answering}

\author{Qingqing Cao, Harsh Trivedi, Aruna Balasubramanian, Niranjan Balasubramanian
\\
Department of Computer Science\\
Stony Brook University\\
Stony Brook, NY 11794, USA \\
\texttt{\{qicao,hjtrivedi,arunab,niranjan\}@cs.stonybrook.edu}
}
\date{}

\begin{document}

\maketitle
%
\begin{abstract}


Transformer-based QA models use input-wide self-attention -- i.e. across both the question and the input passage -- at all layers, causing them to be slow and memory-intensive. It turns out that we can get by without input-wide self-attention at all layers, especially in the lower layers. We introduce \emph{DeFormer}, a decomposed transformer, which substitutes the full self-attention with question-wide and passage-wide self-attentions in the lower layers. This allows for question-independent processing of the input text representations, which in turn enables pre-computing passage representations reducing runtime compute drastically. Furthermore, because \sys\ is largely similar to the original model, we can initialize \sys\ with the pre-training weights of a standard transformer, and directly fine-tune on the target QA dataset. We show \sys\ versions of BERT and XLNet can be used to speed up QA by over 4.3x and with simple distillation-based losses they incur only a 1\% drop in accuracy. We open source the code at \url{https://github.com/StonyBrookNLP/deformer}.

\end{abstract}

\section{Introduction}

There is an increasing need to push question answering (QA) models in large volume web scale services \citep{UnderstandingSearchesBetter2019} and also to push them to resource constrained mobile devices for privacy and other performance reasons \citep{deqa}.
State-of-the-art QA systems, like many other NLP applications, are built using large pre-trained Transformers (e.g., BERT~\cite{bert}, XLNet~\cite{xlnet}, Roberta~\cite{roberta}). However, inference in these models requires prohibitively high-levels of runtime compute and memory making it expensive to support large volume deployments in data centers and infeasible to run on resource constrained mobile devices. 
\begin{figure}[t!]
    \centering
    \includegraphics[width=\linewidth]{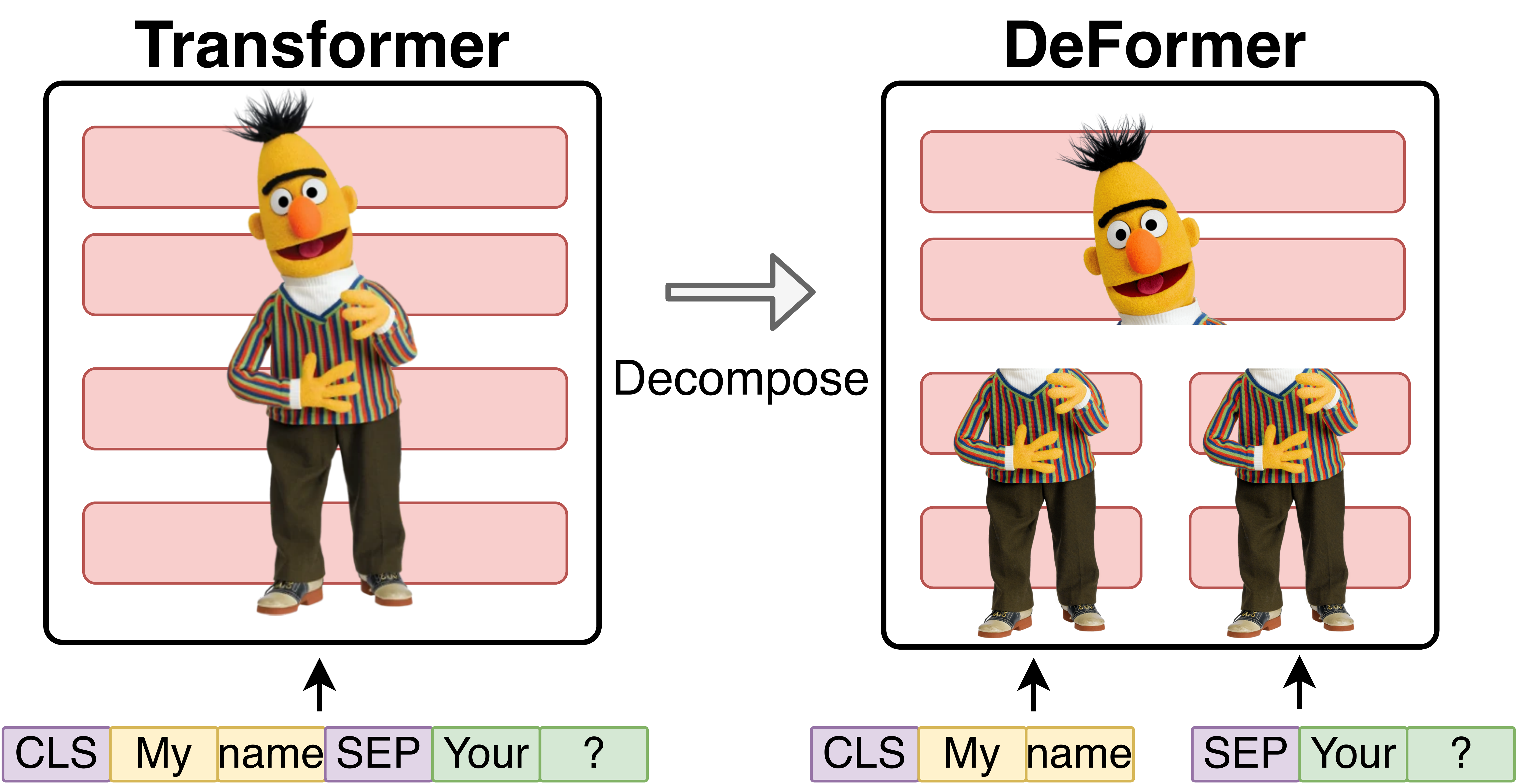}
    \caption{Original Transformer applies full self-attention to encode the concatenated question and passage sequence, while \sys\ encodes the question and passage independently in the lower layers and processes them jointly in the higher layers.}
    \label{fig:deformer}
\end{figure} 

Our goal is to take pre-trained Transformer-based models and modify them to enable faster inference for QA without having to repeat the pre-training. This is a critical requirement if we want to explore many points in the accuracy versus speed trade-off because pre-training is expensive. 

The main compute bottleneck in Transformer-based models is the input-wide self-attention computation at each layer. In reading comprehension style QA, this amounts to computing self-attention over the question and the context text together. This helps the models create highly effective \textit{question-dependent} context representations and vice-versa. Of these, building representations of the context takes more time because it is typically much longer than the question. If the context can be processed independent of the question, then this expensive compute can be pushed \textit{offline} saving significant runtime latency. 

Can we process the context independent of the question, at least in some of the layers, without too much loss in effectiveness? There are two empirical observations that indicate that this is possible. First, previous studies have demonstrated that lower layers tend to focus on local phenomena such as syntactic aspects, while the higher layers focus on global (long distance) phenomena such as semantic aspects relevant for the target task~\citep{bert_nlp_pipeline, hao2019visualizing, Clark_2019}. Second, as we show later (see Section \ref{sec:method}), in a standard BERT-based QA model, there is less variance in the lower layer representations of text when we vary the question. This means that in the lower layers information from the question is not as critical to form text representations. 
Together, these suggest that considering only local context in lower layers of Transformer and considering full global context in upper layers can provide speedup at a very small cost in terms of effectiveness.

Based on these observations, we introduce \sys\, a simple decomposition of pre-trained Transformer-based models, where lower layers in the decomposed model process the question and context text independently and the higher layers process them jointly (see Figure~\ref{fig:deformer} for a schematic illustration). Suppose we allow $k$ lower layers in a $n$-layer model to process the question and context text independently. 
\sys\ processes the context texts through $k$ lower layers offline and caches the output from the $k$-th layer. During runtime the question is first processed through the $k$-layers of the model, and the text representation for the $k$-th layer is loaded from the cache. These two $k$-th layer representations are fed to the $(k+1)$-th layer as input and further processing continues through the higher layers as in the original model. In addition to directly reducing the amount of runtime compute, this also reduces memory significantly as the intermediate text representations for the context are no longer held in memory.

A key strength of this approach is that one can make any pre-trained Transformer-based QA model faster by creating a corresponding \sys\ version that is directly fine-tuned on the target QA datasets without having to repeat the expensive pre-training. Our empirical evaluation on multiple QA datasets show that with direct fine-tuning the decomposed model incurs only a small loss in accuracy compared to the full model.

This loss in accuracy can be reduced further by learning from the original model. We want \sys\ to behave more like the original model. In particular, the upper layers of \sys\ should produce representations that capture the same kinds of information as the corresponding layers in the original model. We add two distillation-like auxiliary losses ~\citep{hinton2015distilling}, which minimize the output-level and the layer-level divergences between the decomposed and original models.

We evaluate \sys\ versions of two transformer-based models, BERT and XLNet on three different QA tasks and two sentence-sentence paired-input tasks\footnote{These simulate other information seeking applications where one input is available offline.}. \sys\ achieves substantial speedup (2.7 to 4.3x) and reduction in memory (65.8\% to 72.9\%) for only small loss in effectiveness (0.6 to 1.8 points) for QA. Moreover, we find that \sys\ version of BERT-large is faster than the original version of the smaller BERT-base model, while still being more accurate. Ablations shows that the supervision strategies we introduce provide valuable accuracy improvements and further analysis illustrate that \sys\ provides good runtime vs accuracy trade-offs.










\section{Decomposing Transformers for Faster Inference}
\label{sec:method}


The standard approach to using transformers for question answering is to compute the self-attention over both question and the input text (typically a passage). This yields highly effective representations of the input pair since often what information to extract from the text depends on the question and vice versa. If we want to reduce complexity, one natural question to ask is whether we can decompose the Transformer function over each segment of the input, trading some representational power for gains in ability to push processing the text segment offline. 

\begin{figure}
	\begin{center}
		\includegraphics[width=\linewidth]{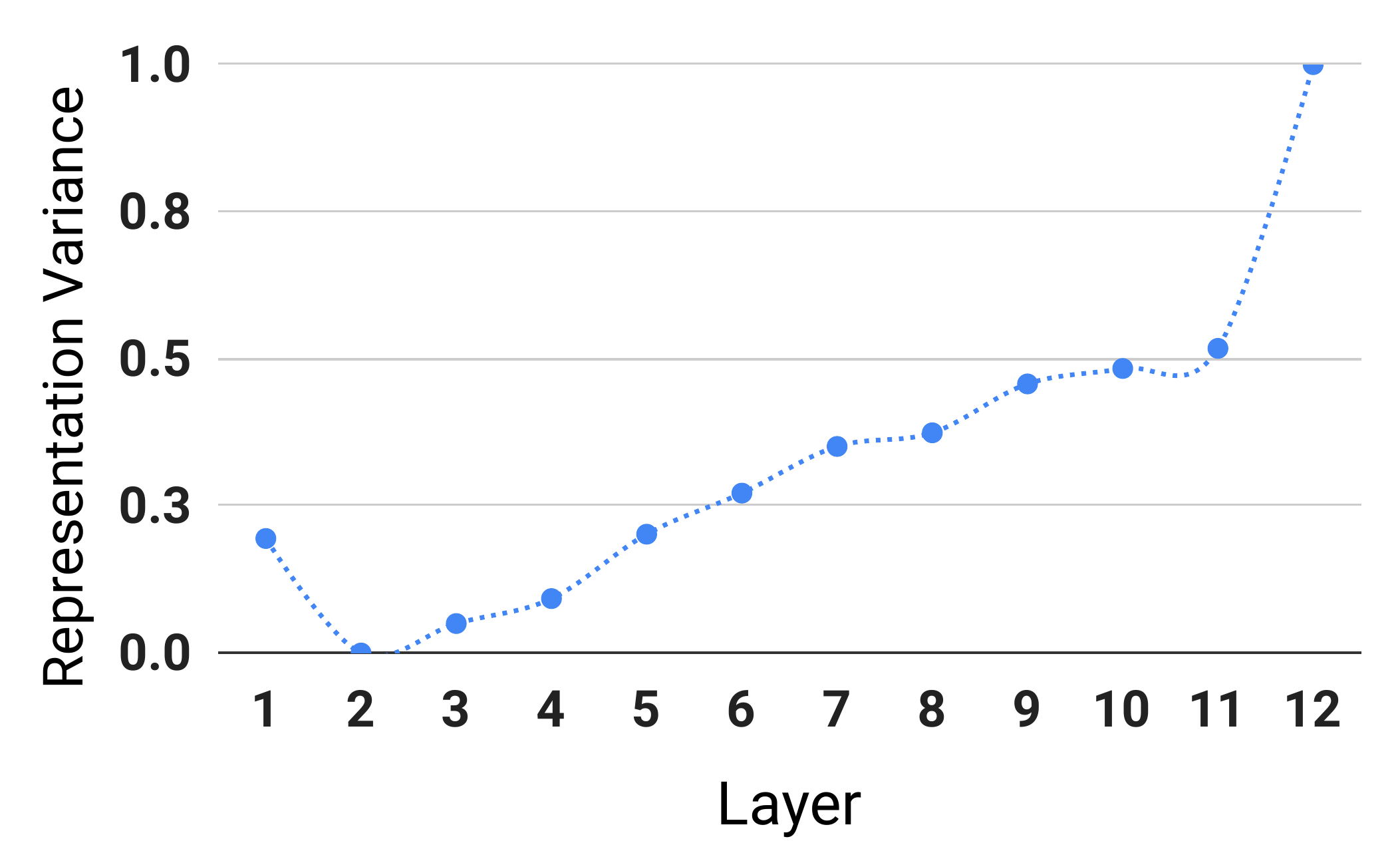}
	\end{center}
	\vspace{-0.5cm}
	\caption{Normalized variance of passage representations when paired with different questions at different layers. We define the representation variance as the average cosine distance from the centroid to all representation vectors. In this figure, the variance is averaged for 100 paragraphs (each paired with 5 different questions) and normalized to [0, 1]. Smaller variance in the lower layers indicates the passage representation depends less on the question, while higher variance in the upper layers shows the passage representation relies more on the interaction with the question.}
	\label{fig:passage_variance}\vspace{-0.1cm}
\end{figure}

The trade-off depends on how important it is to have attention from question tokens when forming text representations (and vice versa) in the lower layers. To assess this, we measured how the text representation changes when paired with different questions. In particular, we computed the average passage representation variance when paired with different questions. The variance is measured using cosine distance between the passage vectors and their centroid. As Figure~\ref{fig:passage_variance} shows that in the lower layers, the text representation does not change as much as it does in the upper layers, suggesting ignoring attention from question tokens in lower layers may not be a bad idea. This is also in agreement with results on probing tasks which suggest that lower layers tend to model mostly local phenomena (e.g., POS, syntactic categories), while higher layers tend to model more semantic phenomena that are task dependent (e.g, entity co-reference) relying on wider contexts. 

\begin{figure*}[t!]
	\begin{center}
		\includegraphics[width=\linewidth]{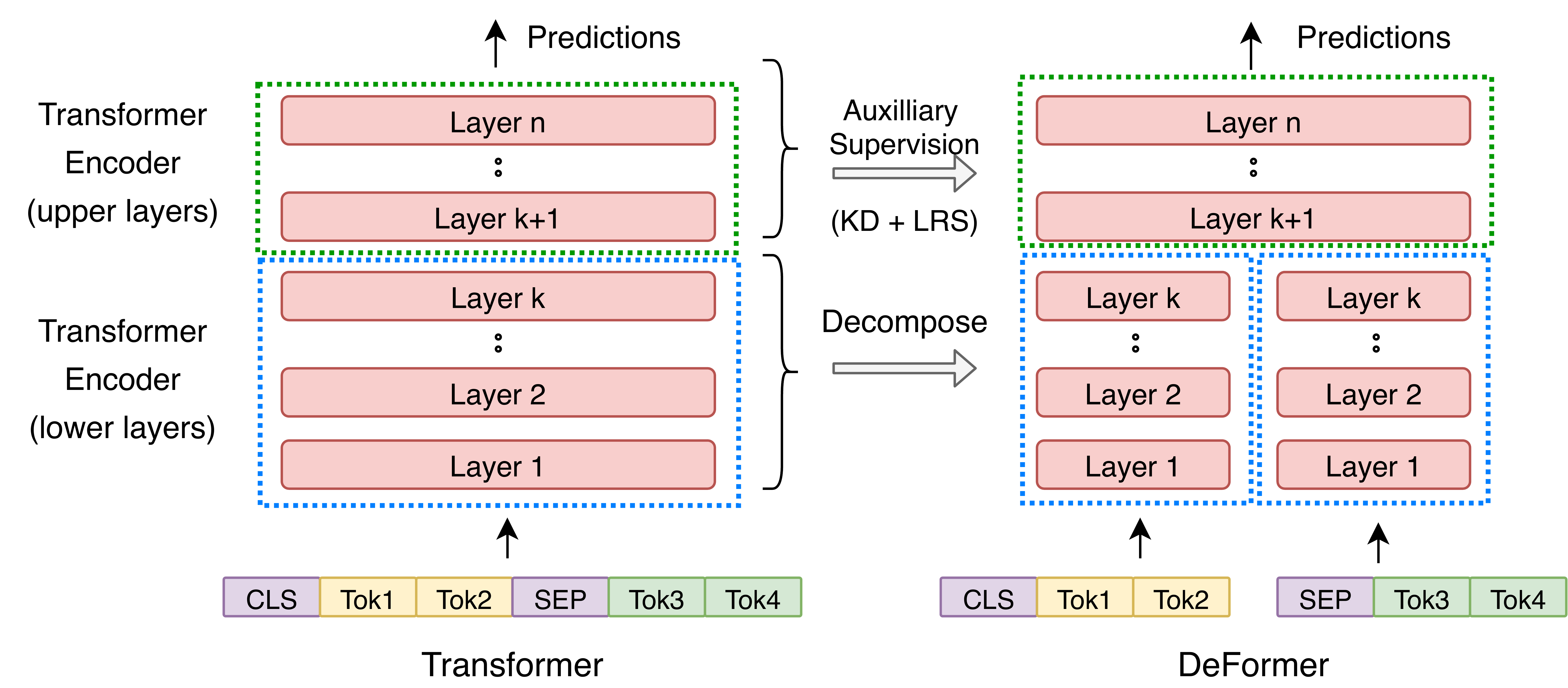}
	\end{center}
	\caption{Decomposing Transformers up to layer $k$, which enables encoding each segment independently from layer $1$ to layer $k$. Auxiliary supervision of upper layer information from the original model further helps the decomposed model to compensate for information loss in the lower layers. KD is Knowledge Distillation loss and LRS is Layerwise Representation Similarity loss.}
	\label{fig:idea}
\end{figure*}

Here we formally describe our approach for decomposing attention in the lower layers to allow question independent processing of the contexts. 

\subsection{\sys}
First, we formally define the computation of a Transformer for a paired-task containing two segments of text, $T_a$ and $T_b$. Let the token embedding representations of segment $T_a$ be $\mathbf{A}=[\mathbf{a_1} ; \mathbf{a_2}; ...; \mathbf{a_{q}}]$ and of $T_b$ be $\mathbf{B}=[\mathbf{b_1} ; \mathbf{b_2}; ...; \mathbf{b_{p}}]$. The full input sequence $\mathbf{X}$ can be expressed by concatenating the token representations from segment $T_a$ and $T_b$ as $X = [A ; B]$. The Transformer encoder has $n$ layers (denoted $L_i$ for layer i), which transform this input sequentially: $X^{l+1} = L_i(X^{l})$. For the details of the Transformer layer, we refer the reader to \citep{self-attn}. We denote the application of a stack of layers from layer i to layer j be denoted as $L_{i:j}$. The output representations of the full Transformer, $\mathbf{A^{n}}$ and $\mathbf{B^{n}}$ can be written as:

\begin{equation}
    [\mathbf{A^{n}}; \mathbf{B^{n}}] = L_{1:n}([\mathbf{A^{0}}; \mathbf{B^{0}}])
\end{equation}


Figure \ref{fig:idea} shows a schematic of our model. We decompose the computation of lower layers (up to layer $k$) by simply removing the cross-interactions between $T_a$ and $T_b$ representations. Here $k$ is a hyper-parameter. The output representations of the decomposed Transformer, $\mathbf{A^{n}}$ and $\mathbf{B^{n}}$ can be expressed as:

\begin{equation}
    [\mathbf{A^{n}}; \mathbf{B^{n}}] = L_{k+1:n}([L_{1:k}(\mathbf{A^{0}});L_{1:k}(\mathbf{B^{0}}))
\end{equation}

Transformer-based QA systems process the input question and context together through a stack of self-attention layers. So applying this decomposition to Transformer for QA allows us to process the question and the context text independently, which in turn allows us to compute the context text representations for lower layers offline. With this change the runtime complexity of each lower layer is reduced from $O((p+q)^2)$ to $O(q^2 + c)$, where $c$ denotes cost of loading the cached representation.


\subsection{Auxiliary Supervision for \sys}

\sys\ can be used in the same way as the original Transformer. Since \sys\ retains much of the original structure, we can initialize this model with the pre-trained weights of the original Transformer and fine-tune directly on downstream tasks. 
However, \sys\ looses some information in the representations of the lower layers. The upper layers can learn to compensate for this during fine-tuning. However, we can go further and use the original model behavior as an additional source of supervision.


Towards this end, we first initialize the parameters of \sys\ with the parameters of a pre-trained full Transformer, and fine-tune it on the downstream tasks. We also add auxiliary losses that make \sys\ predictions and its upper layer representations closer to the predictions and corresponding layer representations of the full Transformer.

\textbf{Knowledge Distillation Loss:} 
We want the prediction distribution of \sys\ to be closer to that of the full Transformer. We minimize the Kullback---Leibler divergence between decomposed Transformer prediction distribution $P_A$ and full Transformer prediction distribution $P_B$:

\begin{equation*}
    \mathcal{L}_{kd} = D_{KL}(P_A \| P_B)
\end{equation*}

\textbf{Layerwise Representation Similarity Loss:} We want the upper layer representations of \sys\ to be closer to those of full Transformer. We minimize the euclidean distance between token representations of the upper layers of decomposed Transformer and the full Transformer. Let $\mathbf{v}_i^{j}$ be the representation of the $j^{th}$ token in the $i^{th}$ layer in the full transformer, and let $\mathbf{u}_i^{j}$  be the corresponding representation in \sys. For each of the upper layers $k+1$ through $n$, we compute a layerwise representation similarity (lrs) loss as follows:

\begin{equation*}
    \mathcal{L}_{lrs} = \sum_{i=k}^{n}\sum_{j=1}^{m} \| \mathbf{v}_j^{i} - \mathbf{u}_j^{i} \|^2
\end{equation*}

We add the knowledge distillation loss ($\mathcal{L}_{kd}$) and layerwise representation similarity loss ($\mathcal{L}_{lrs}$) along with the task specific supervision Loss ($\mathcal{L}_{ts}$) and learn their relative importance via hyper-parameter tuning:

\begin{equation}
    \mathcal{L}_{\text{total}} = \gamma \mathcal{L}_{ts} + \alpha \mathcal{L}_{kd} + \beta \mathcal{L}_{lrs}
\end{equation}

We use Bayesian Optimization ~\citep{bayesopt} to tune the $\gamma$, $\alpha$ and $\beta$ instead of simple trial-and-error or grid/random search. This is aimed at reducing the number of steps required to find a combination of hyper-parameters that are close to the optimal one.

\section{Evaluation}


\subsection{Datasets}

We use the pre-trained uncased BERT base and large\footnote{Whole Word Masking version} models on five different {\em paired-input} problems covering 3 QA tasks, and in addition two other sentence-sentence tasks\footnote{We pick these as additional datasets to show the utility of decomposition in other information seeking applications similar to QA, where one of the inputs can be assumed to be available offline. For instance, we may want to find answer (premise) sentences from a collection that support information
contained in a query (hypothesis) sentence. Another use case is FAQ retrieval, where a user question is compared against a collection of previously asked
questions.
}.  

\noindent{\textbf{SQuAD v1.1} (Stanford Question Answering Dataset)}~\citep{rajpurkar2016squad} is an extractive question answering datasets containing \textgreater 100,000 question and answer pairs generated by crowd workers on Wikipedia articles. 

\noindent{\textbf{RACE}}  ~\citep{lai2017race} is reading comprehension dataset collected from the English exams that are designed to evaluate the reading and reasoning ability of middle and high school Chinese students. It has over 28,000 passages and 100,000$+$ questions.

\noindent{\textbf{BoolQ}} ~\citep{clark2019boolq} consists of 15942 yes/no questions that are naturally occurring in unprompted and unconstrained settings. 

\noindent{\textbf{MNLI} (Multi-Genre Natural Language Inference)} ~\citep{multinli}  is a crowd-sourced corpus of 433k sentence pairs annotated with textual entailment information.

\noindent{\textbf{QQP} (Quora Question Pairs)}~\citep{qqp} consists of over 400,000 potential duplicate question pairs from Quora.

For all 5 tasks, we use the standard splits provided with the datasets but in addition divide the original training data further to obtain a 10\% split to use for tuning hyper-parameters (tune split), and use the original development split for reporting efficiency (FLOPs, memory usage) and effectiveness metrics (accuracy or F1 depending on the task).

\subsection{Implementation Details}

We implement all models in TensorFlow 1.15 ~\citep{tensorflow} based on the original BERT  ~\citep{bert} and the XLNet~\citep{xlnet} codebases. We perform all experiments on one TPU v3-8 node (8 cores, 128GB memory) with \textit{bfloat16} format enabled. We measure the FLOPs and memory consumption through the TensorFlow Profiler\footnote{\url{https://www.tensorflow.org/versions/r1.15/api_docs/python/tf/profiler/profile}}. For \sys\ models, we tune the hyperparameters for weighting different losses using bayesian optimizaiton libray~\citep{bayesopt_lib} with 50 iterations on the tune split (10\% of the original training sets) and report the performance numbers on the original dev sets. The search range is [0.1, 2.0] for the 3 hyper-parameters. We put the detail hyper-parameters in the section \ref{sec:appendix}.

For \sys-BERT and \sys-XLNet, we compute the representations for one of the input segments offline and cache it. For QA we cache the passages, for natural language inference, we cache the premise\footnote{One use case is where we want to find (premise) sentences from a collection that support information contained in a query (hypothesis) sentence.} and for question similarity we cache the first question\footnote{One use case is FAQ retrieval, where a user question is compared against a collection of previously asked questions}.



\subsection{Results}
Table \ref{table:main-results} shows the main results comparing performance, inference speed and memory requirements of BERT-base and \sys-BERT-base when using nine lower layers, and three upper layers (see Subsection ~\ref{subsec:ablation} for the impact of the choice of upper/lower splits). We observe a substantial speedup and significant memory reduction in all the datasets, while retaining most of the original model's effectiveness (as much as 98.4\% on SQuAD and 99.8\% on QQP datasets), the results of XLNet in the same table demonstrates the decomposition effectiveness for different pre-trained Transformer architectures. Table \ref{table:pairwise-results} shows that the decomposition brings 2x speedup in inference and more than half of memory reduction on both QQP and MNLI datasets, which take pairwise input sequences. The effectiveness of decomposition generalizes further beyond QA tasks as long as the input sequences are paired. Efficiency improvements increase with the size of the text segment that can be cached.


\begin{table*}[ht!]
    \centering
    \setlength\tabcolsep{7pt}
    \small
    \begin{tabular}{cccccccc}\toprule
       Model       &   Datasets      & Avg. Input& Original & \sys-   &  Performance Drop     & Inference  & Memory \\
              &               & Tokens    & base &  base &  (absolute $|$ \%age) & Speedup    & Reduction \\
              &               &           &      &           &                       & (times)    & (\%age) \\
        \midrule
      & SQuAD                & 320 & 88.5 & 87.1 & 1.4\thinspace$|$\thinspace1.6 & 3.2x & 70.3 \\
 BERT & RACE                 & 2048 & 66.3 & 64.5 & 1.8\thinspace$|$\thinspace2.7 & 3.4x & 72.9 \\
      & BoolQ                & 320 & 77.8 & 76.8 & 1.0\thinspace$|$\thinspace1.3 & 3.5x & 72.0 \\
        \midrule
      & SQuAD                & 320 & 91.6 & 90.4 & 1.2\thinspace$|$\thinspace1.3 & 2.7x & 65.8 \\
XLNet & RACE                 & 2048 & 70.3 & 68.7 & 1.6 \thinspace$|$\thinspace2.2 & 2.8x & 67.6 \\
      & BoolQ                 & 320 & 80.4 & 78.8 & 0.6 \thinspace$|$\thinspace0.7 & 3.0x & 68.3 \\
        \bottomrule
    \end{tabular}
    \caption{(i) Performance of original fine-tuned vs fine-tuned models of \sys-BERT-base and \sys-XLNet-base, (ii) Performance drop, inference speedup and inference memory reduction of \sys- over original models for 3 QA tasks. \sys-BERT-base uses nine lower layers, and three upper layers with caching enabled, \sys-XLNet-base use eight lower layers, and four upper layers with caching enabled. For SQuAD and RACE we also train with the auxiliary losses, and for the others we use the main supervision loss -- the settings that give the best effectiveness during training. Note that the choice of the loss doesn't affect the efficiency metrics.
    }
    \label{table:main-results}

\end{table*}

\begin{table*}[t!]
    \centering
    \setlength\tabcolsep{7pt}
    \small
    \begin{tabular}{lcccccc}\toprule
                             & Avg. Input& BERT & \sys-   &  Performance Drop     & Inference  & Memory \\
                             & Tokens    & base & BERT base &  (absolute $|$ \%age) & Speedup    & Reduction \\
                             &           &      &           &                       & (times)    & (\%age) \\
        \midrule
        MNLI                 & 120 & 84.4 & 82.6 & 1.8\thinspace$|$\thinspace2.1 & 2.2x & 56.4 \\
        QQP                  & 100 & 90.5 & 90.3 & 0.2\thinspace$|$\thinspace0.2 & 2.0x & 50.0 \\
        \bottomrule
    \end{tabular}
    \caption{(i) Performance of BERT-base vs \sys-BERT-base, (ii) Performance drop, inference speedup and inference memory reduction of \sys-BERT-base over BERT-base for 2 pairwise tasks. \sys-BERT-base uses nine lower layers, and three upper layers with caching enabled. }
    \label{table:pairwise-results}
\end{table*}

\textbf{Small Distilled or Large Decomposed?} Table \ref{table:main-results2} compares performance, speed and memory of BERT-base, BERT-large and \sys-BERT-large. \sys-BERT-large is 1.6 times faster than the smaller BERT-base model. Decomposing the larger model turns out to be also more effective than using the smaller base model (+2.3 points) This shows that with decomposition, a large Transformer can run faster than a smaller one which is half its size, while also being more accurate. 

Distilling a larger model into a smaller one can yield better accuracy than training a smaller model from scratch. As far as we know, there are two related but not fully comparable results. (1) \citet{Tang2019DistillingTK} distill BERT to a small LSTM based model where they achieve 15x speedup but at a significant drop in accuracy of more than 13 points on MNLI. (2) \citet{distilbert} distill BERT to a smaller six layer Transformer, which can provide 1.6x speedup but gives \textgreater2 points accuracy drop on MNLI and \textgreater3 points F1 drop on SQuAD. A fair comparison requires more careful experimentation exploring different distillation sizes which requires repeating pre-training or data augmentation -- an expensive proposition.





\begin{table*}[t!]
    \centering
    \setlength\tabcolsep{7pt}
    \small
    \begin{tabular}{lccccc}\toprule
                             & Performance (Squad-F1) & Speed (GFLOPs) & Memory (MB) \\
        \midrule
        BERT-large           & 92.3 & 204.1 & 1549.6 \\
        BERT-base            & 88.5 & 58.4  & 584.2 \\
        \sys-BERT-large    & 90.8 & 47.7  & 359.7 \\
        \bottomrule
    \end{tabular}
    \caption{Performance, Inference Speed and Memory for different models on SQuAD.}
    \label{table:main-results2}
\end{table*}

{\bf Device Results:} To evaluate the impact on different devices, we deployed the models on three different machines (a GPU, CPU, and a mobile phone). Table \ref{table:latency} shows the average latency in answering a question measured on a subset of the SQuAD dataset. On all devices, we get more than three times speedup.

\begin{table}[htb!]
    \setlength\tabcolsep{3pt}
    \begin{tabular}{lcc}\toprule
    & {BERT} & {\sys-BERT} \\ \midrule
    Tesla V100 GPU       & 0.22                          & 0.07                               \\
    Intel i9-7900X CPU   & 5.90                          & 1.66                               \\
    OnePlus 6 Phone      & 10.20*                        & 3.28*                              \\
    \bottomrule
    \end{tabular}
    \caption{Inference latency (in seconds) on SQuAD datasets for BERT-base vs \sys-BERT-base, as an average measured in batch mode. On the GPU and CPU batch size is 32 and on the phone (marked by *) batch size is 1.}
    \label{table:latency}
\end{table}


\subsection{Ablation Study}
\label{subsec:ablation}

Table \ref{table:ablation} shows the contribution of auxiliary losses for fine-tuning \sys-BERT on SQuAD dataset. The drop in effectiveness when not using Layerwise Representation Similarity (LRS in table), and Knowlege Distillation (KD) losses shows the utility of auxiliary supervision.  

\begin{table}[htb!]
    \centering
    \setlength\tabcolsep{3pt}
    \small
    \begin{tabular}{lccccc}\toprule
                                     & Base Model & Large Model \\
        \midrule
        BERT                         & 88.5 & 92.3 \\
        \midrule
        \sys-BERT                  & 87.1 & 90.8 \\
        \qquad w/o LRS        & 86.2 & 88.9 \\
        \qquad w/o KD  \& LRS & 85.8 & 87.5 \\
        \bottomrule
    \end{tabular}
    \caption{Ablation analysis on SQuAD datasets for \sys-BERT-base and \sys-BERT-large models. LRS is the layerwise representation similarity loss. KD is the knowledge distillation loss on the prediction distributions.}
    \label{table:ablation}
\end{table}

\begin{figure*}[ht]
	\centering{
	\begin{subfigure}{0.465\textwidth}
		\includegraphics[width=\textwidth]{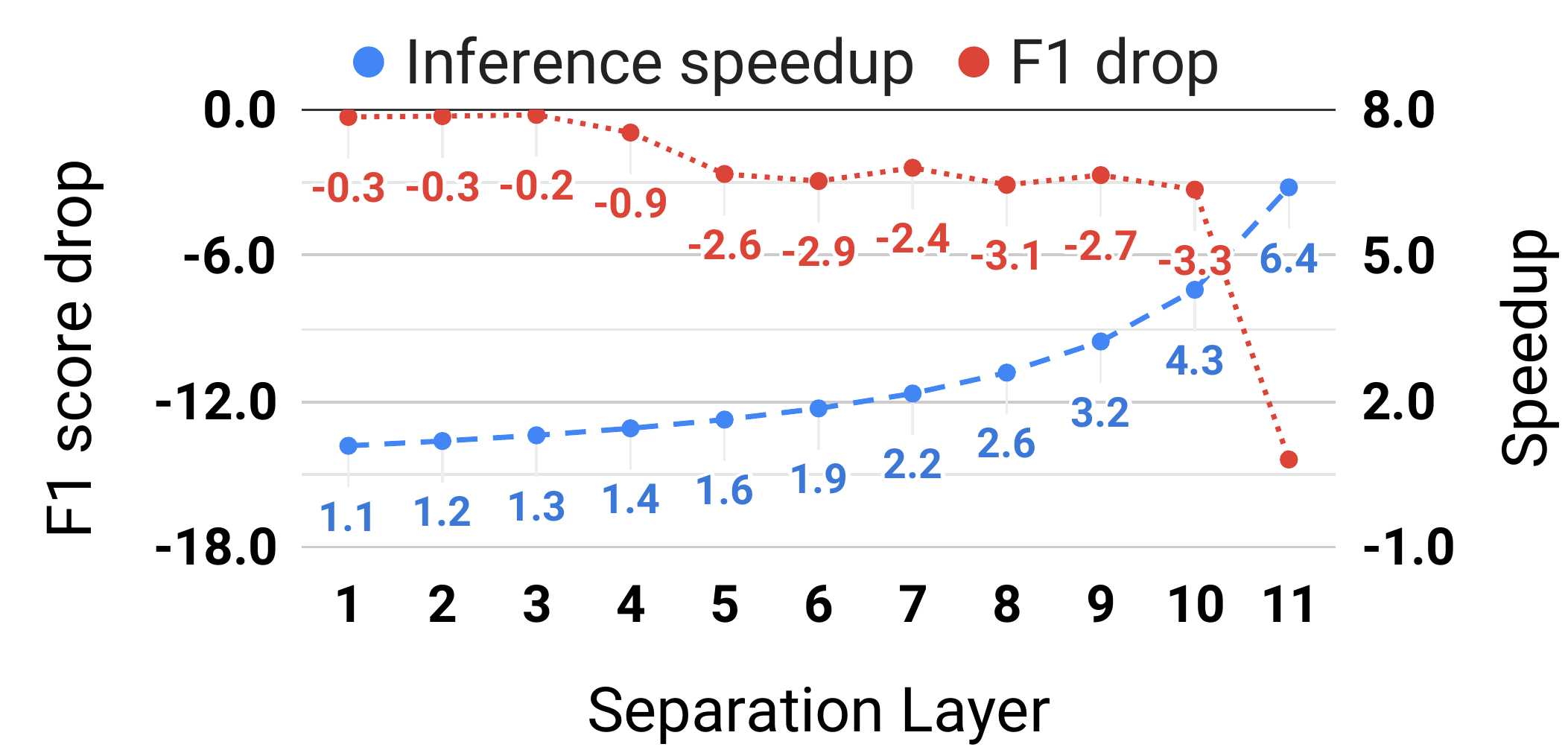}
		\caption{F1 drop versus speedup on SQuAD for \sys-BERT-base without auxiliary supervision.}
		\label{fig:f1_speedup_base}
	\end{subfigure}
	\quad
	\begin{subfigure}{0.465\textwidth}
		\includegraphics[width=\textwidth]{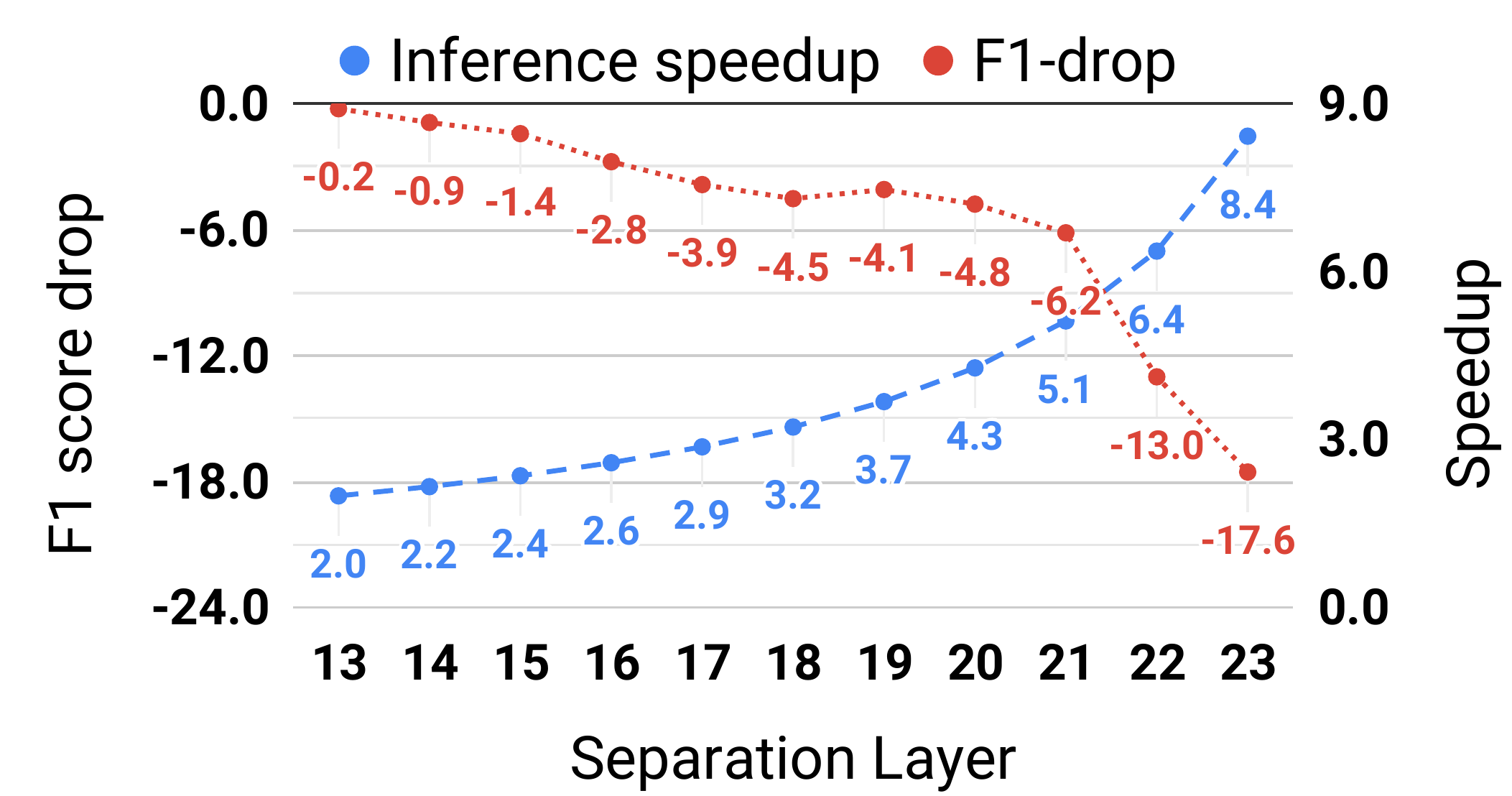}
		\caption{F1 drop versus speedup on SQuAD for \sys-BERT-large without auxiliary supervision.}
		\label{fig:f1_speedup_large}
	\end{subfigure}
	\caption{F1 drop versus speedup of \sys-BERT model (without auxiliary supervision) when separating at different layers.}
	}
\end{figure*}

Figure \ref{fig:f1_speedup_base} and figure \ref{fig:f1_speedup_large} show how the effectiveness and inference speed of \sys-BERT changes as we change the separation layer. Inference speedup scales roughly quadratically with respect to the number of layers with decomposed attention. The drop in effectiveness, on the other hand, is negligible for separating at lower layers (until layer 3 for the base model and until layer 13 for the large model) and increases slowly after that with a dramatic increase in the last layers closest to the output. The separation layer choice thus allows trading effectiveness for inference speed.

\section{Analyses}

\subsection{Divergence of \sys\ and original BERT representations}
\label{sec:analysis}
The main difference between the original BERT and the \sys-BERT is the absence of cross attention in the lower layers. We analyze the differences between the representations of the two models across all layers. To this end, we randomly select 100 passages from SQuAD dev dataset as well as randomly selecting 5 different questions that already exist in the dataset associated with each passage. For each passage, we encode all 5 question-passage pair sequence using both the fine-tuned original BERT-base model and the \sys-BERT-base model, and compute their distance of the vector representations at each layer. 

\begin{figure*}[!ht]
	\centering{
		\begin{subfigure}{0.465\textwidth}
			\includegraphics[width=\textwidth]{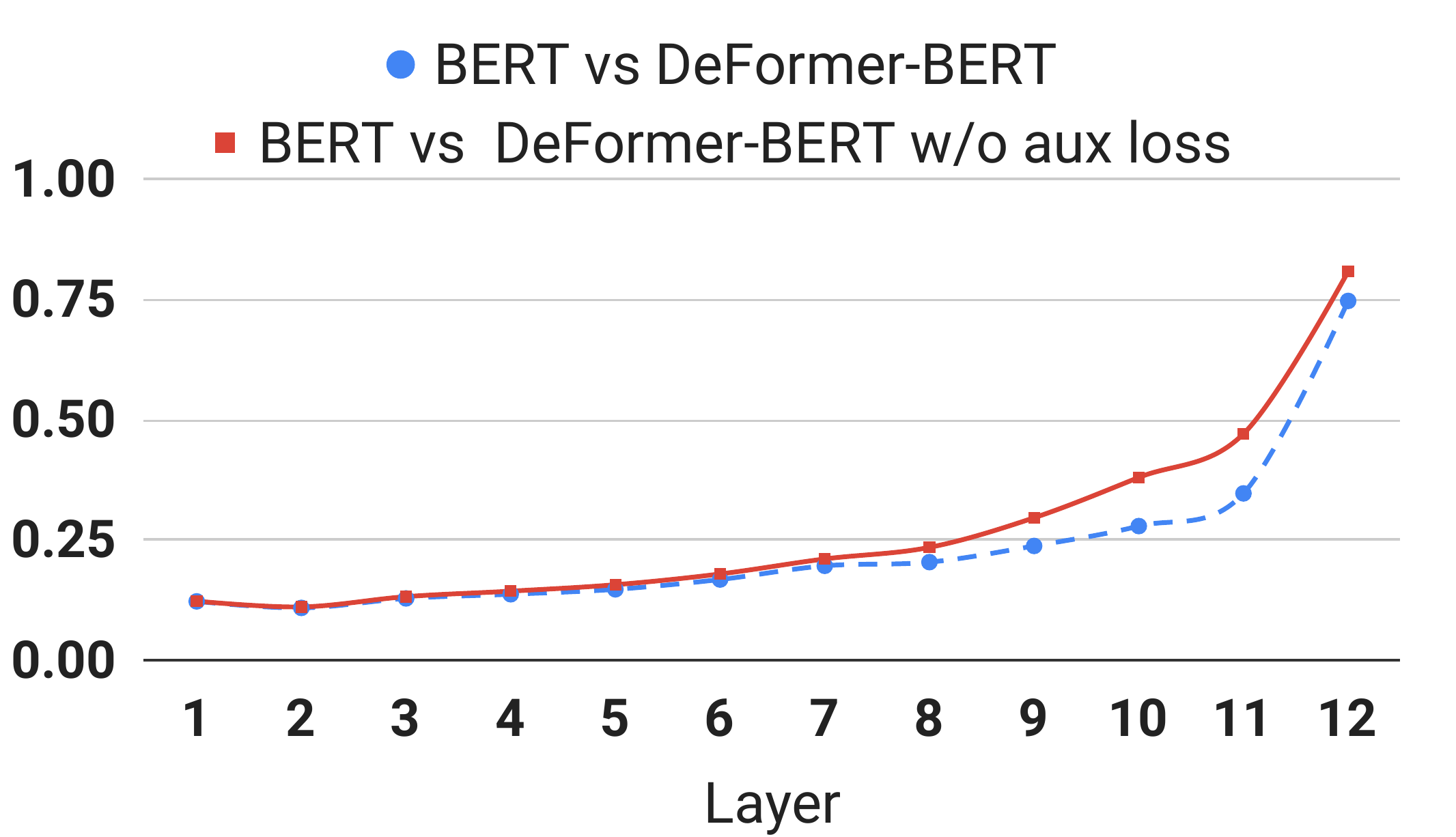}
			\caption{Passage distance comparison}
			\label{fig:passage_distance_compare}
		\end{subfigure}
		\quad
		\begin{subfigure}{0.465\textwidth}
			\includegraphics[width=\textwidth]{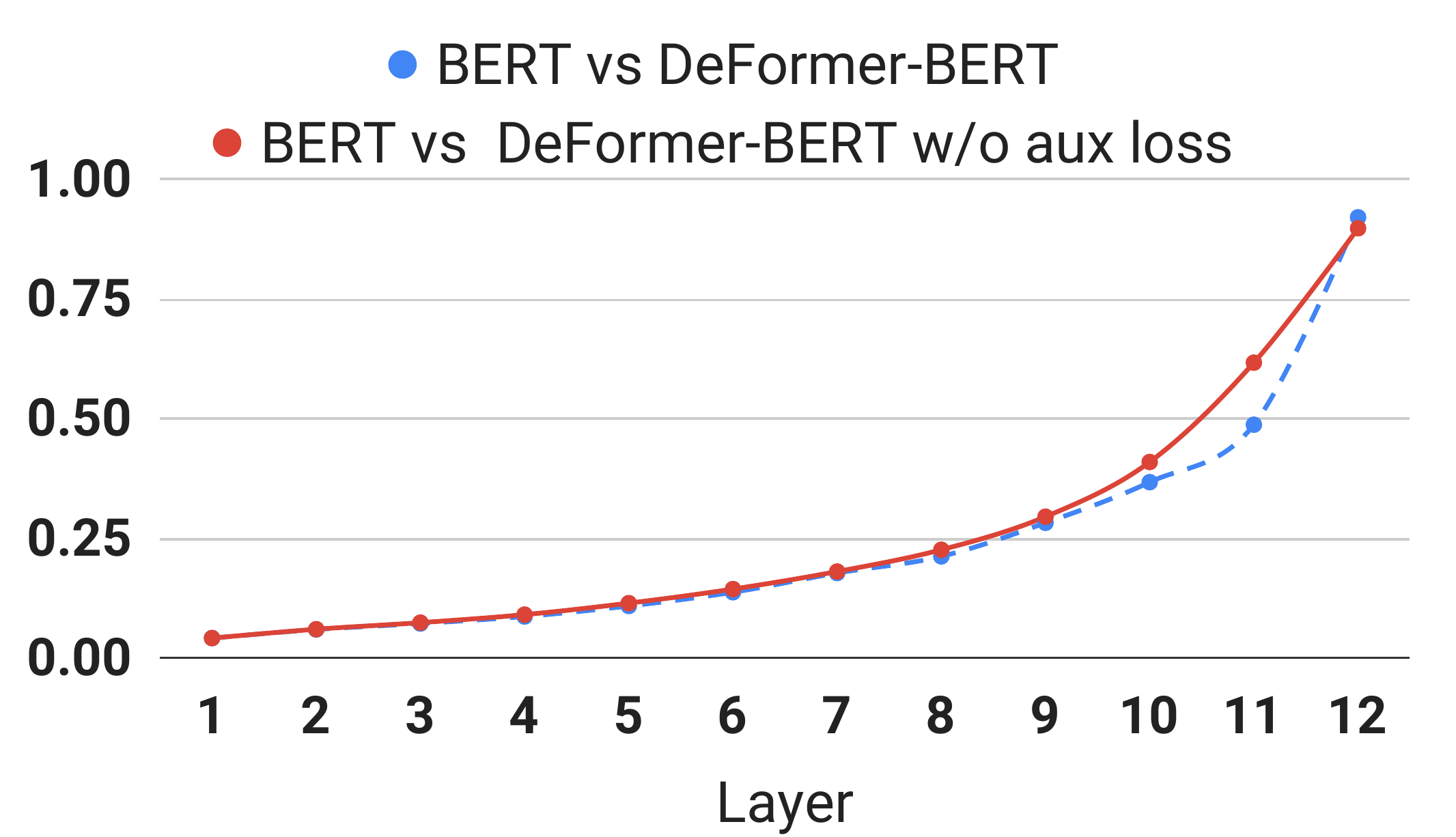}
			\caption{Question distance comparison}
			\label{fig:question_distance_compare}
		\end{subfigure}
		\caption{Representation distance of BERT vs \sys-BERT and distance of BERT vs \sys-BERT w/o auxiliary loss/supervision}\label{fig:sequence_distance_cmp}
	}
\end{figure*}

Figure \ref{fig:sequence_distance_cmp} shows the averaged distances of both the question and passage at different layers. The lower layer representations of the passage and questions for both models remain similar but the upper layer representations differ significantly, supporting the idea that lack of cross-attention has less impact in the lower layers than in the higher ones. Also, using the auxiliary supervision of upper layers has the desired effect of forcing \sys\ to produce representations that are closer to the original model. This effect is less pronounced for the question representations. 



\subsection{Inference Cost}
\label{sec:inference_cost}

\sys\ enables caching of text representations that can be computed offline. While a full-scale analysis of the detailed trade-offs in storage versus latency is beyond the scope of this paper, we present a set of basic calculations to illustrate that the storage cost of caching can be substantially smaller compared to the inference cost.  
Assuming a use case of evaluating one million question-passage pairs daily, we first compute the storage requirements of the representations of these passages. With the BERT-base representations we estimate this to be 226KB per passage and 226GB in total for 1 million passages. The cost of storing this data and the added compute costs and reading these passages at the current vendor rates amounts to a total of \$61.7 dollars per month. To estimate inference cost, we use the compute times we obtain from our calculations and use current vendor rates for GPU workloads which amounts to \$148.5 dollars to support the 1 million question-passage pair workload. The substantial reduction in cost is because the storage cost is many orders of magnitude cheaper than using GPUs. Details of these calculations are listed in the Appendix.



\section{Related work}

Speeding up inference in a model requires reducing the amount of compute involved. There are two broad related directions of prior work: 

(i) {\bf Compression techniques} can be used to reduce model size through low rank approximation ~\citep{low-rank-approx,kim2015compression,tai2015convolutional,chen2018groupreduce}, and model weights pruning ~\citep{guo2016dynamic,han2015deep}, which have been shown to help speedup inference in CNN and RNN based models. For Transformers, ~\citet{prune-heads} explore pruning the attention heads to gain inference speedup. This is an orthogonal approach that can be combined with our decomposition idea. However, for the paired-input tasks we consider, pruning heads only provides limited speedup. In more recent work ~\citet{btd} propose approximating the quadratic attention computation with a tensor decomposition based multi-linear attention model. However, it is not clear how this multi-linear approximation can be applied to pre-trained Transformers like BERT.

(ii) {\bf Distillation techniques} can be used to train smaller student networks to speedup inference. \citet{Tang2019DistillingTK} show that BERT can be used to guide designing smaller models (such as single-layer BiLSTM) for multiple tasks. But for the tasks we study, such very small models suffer a significant performance drop. For instance there is a 13\% accuracy degration on MNLI task. Another closely related recent work is DistillBERT~\citep{distilbert}, which trains a smaller BERT model (half the size of BERT-base) that runs 1.5 times faster than the original BERT-base.However, the distilled model incurs a significant drop in accuracy. While more recent distillation works such as \citep{tinybert} and \citep{mobilebert} further improve the speedups, our decomposition also achieves similar accuracy performance. More importantly, this distillation model usually undergo expensive pre-training on the language modeling tasks before they can be fine-tuned for the downstream tasks. 

Previous QA neural models like BIDAF\cite{seo2016bidirectional}, QANet\cite{yu2018qanet} and many others contain decomposition as part of their \textit{neural architecture design}. In contrast, the focus of our work is to show that large \textit{pre-trained} Transformer models can be decomposed at the \textit{fine-tuning} stage to bring effectiveness of SOTA pre-trained transformers at much lower inference latency.

In this work, we ask if can we speedup the inference of Transformer models without compressing or removing model parameters. Part of the massive success of pre-trained Transformer models for many NLP task is due to a large amount of parameters capacity to enable complex language representations. The decomposition we propose makes minimal changes retaining the overall capacity and structure of the original model but allows for faster inference by enabling parallel processing and caching of segments.

\sys\ applies to settings where the underlying model relies on input-wide self-attention layers. Even with models that propose alternate ways to improve efficiency, as long as the models use input-wide self-attention, \sys\ can be applied as a complementary mechanism to further improve inference efficiency. We leave an evaluation of applying \sys\ on top of other recent efficiency optimized models for future work.




\section{Conclusion}
Transformers have improved the effectiveness of NLP tools by their ability to incorporate large contexts effectively in multiple layers. This however imposes a significant complexity cost. In this work, we showed that modeling such large contexts may not always be necessary. We build a decomposition of the transformer model that provides substantial improvements in inference speed, memory reduction, while retaining most of the original model's accuracy. A key benefit of the model is that its architecture remains largely the same as the original model which allows us to avoid repeating pre-training and use the original model weights for fine-tuning. The distillation techniques further reduce the performance gap with respect to the original model. This decomposition model provides a simple yet strong starting point for efficient QA models as NLP moves towards increasingly larger models handling wider contexts.

\section*{Acknowledgement}
We thank Google for supporting this research through the Google Cloud Platform credits.

\bibliography{ref}
\bibliographystyle{acl_natbib}

\clearpage
\appendix

\section{Appendix}
\label{sec:appendix}

Data centers often use GPUs for inference workloads ~\citep{gpu-data-center}, we use the GPUs by default for both models. We use $g_u$ to denote the cost of using one GPU per hour,
$n_{seq}$ to stand for the number of input sequences to process, $b$ for the GPU batch size, and $t_{b}$ is the time (in seconds) take to process $b$ sequences, $s$ denotes the storage size of the cached representations, $s_u$ denotes the cost of storage per month, $r_u$ is the cost of performing 10,000 reading operations (such as loading cached representations from the disk).

The total cost of the original model $Cost_{original}$ is the cost of using GPUs and is given by the formula as below:

$$ Cost_{original}= t_{b} \cdot  \frac{n_{seq}}{b} \cdot \frac{g_u}{3600} $$

And the total cost of the decomposed model $Cost_{decomp}$ includes three parts: using GPUs, storing representations on disk and loading them into memory. It is formulated as:

\begin{multline}
\nonumber Cost_{decomp}= t_{b} \cdot  \frac{n_{seq}}{b} \cdot \frac{g_u}{3600} + \frac{n_{seq}}{b} \cdot \frac{r_u}{10,000} \\ 
+ \frac{s \cdot s_u}{30*24*3600}
\end{multline}

We assume a passage has 150 tokens on average (The number is calculated based on the SQuAD dataset).

We take one cloud service provider ~\citep{cloud-price} to instantiate $g_u$, $s_u$, and $r_u$: one Tesla V100 GPU (16GB memory) costs \$2.48 USD per hour ($g_u=2.48$), 1GB storage takes \$0.02 per month ($s_u=0.02$) and additional \$0.004 per 10,000 read operations ($r_u=0.004$)\footnote{Class B operations on GCP}. 

It takes 226KB to store the vectors for 150 tokens \footnote{vector dimension=768, bfloat16 format}, and the total storage for 1 million sequences is 226GB. The Tesla V100 GPU allows a maximum batch size of 640\footnote{\textgreater 640 batch size will cause V100 GPU out of memory}. We measure the $t_b=4.6$ for the original BERT-base model and $t_b=1.4$ for the decomposed BERT-base model. Then $Cost_{original}=30* 4.6 * 1,000,000 / 640 * 2.48 /3600=\$148.5$, and $Cost_{decomp}= 30* 1.4 * 1,000,000 / 640 * 2.48 /3600 + 30*1,000,000 /10,000 * 0.004 + 226 *0.02  = \$61.7$. 

\paragraph{Hyper-parameters}
We set the final $\alpha=1.1$, $\beta=0.5$ and $\gamma=0.7$ for supervising BERT-base model on the SQuAD dataset, $\alpha=0.4$, $\beta=0.4$ and $\gamma=0.7$ and on the RACE dataset. For XLNet, we find that simple default parameters ($\alpha=1.1$, $\beta=0.5$ and $\gamma=0.7$) work well for both SQuAD and BoolQ datasets.

\end{document}